\documentclass[conference]{IEEEtran} 
\IEEEoverridecommandlockouts
\usepackage{comment}
\usepackage{balance}
\usepackage{amsmath,amssymb,amsfonts}
\usepackage{algorithmic}
\usepackage{graphicx} 
\usepackage{textcomp}
\usepackage{xcolor}
\usepackage[center]{caption}
\usepackage[utf8]{inputenc}
\usepackage{graphicx}
\usepackage{ragged2e}
\usepackage{array}
\usepackage{siunitx}
\usepackage{amsmath}
\usepackage{multirow}
\usepackage{tcolorbox}
\usepackage{booktabs}
\usepackage{float}
\usepackage{stfloats}
\usepackage{enumitem}
\usepackage{cite}

\graphicspath{ {./images/} }
\def\BibTeX{{\rm B\kern-.05em{\sc i\kern-.025em b}\kern-.08em
    T\kern-.1667em\lower.7ex\hbox{E}\kern-.125emX}}
\usepackage{amsmath,amssymb}

\usepackage{makecell}

\newcommand{\etal}[0]{\textit{et al. }}

\begin{document}

\title{EEG-FuseFormer: A Transformer-Driven Feature Fusion Framework for Seizure Onset Prediction
\thanks{ * Vigneshwar Hariharan and Chithra Reghuvaran are co-first authors.}
}



\author{
  \IEEEauthorblockN{Vigneshwar Hariharan$^{1*}$,
                    Chithra Reghuvaran$^{2*}$,
                    Arlene John$^{3}$,
                    Nhat Pham$^{4}$,\\
                    Omer Rana$^{4}$,
                    Deepu John$^{2}$,
                    Ganesh Neelakanta Iyer$^{1}$}
                    \textit{
      $^{1}$National University of Singapore,
      $^{2}$University College Dublin,
      $^{3}$University of Twente,
      $^{4}$Cardiff University
} 
}

\maketitle

\begin{abstract}
Epilepsy is one of the most common neurological disorders globally, characterized by recurring seizures and significantly impacting the quality of life. Despite advancements in diagnostic techniques, the mitigation of risks faced by epilepsy patients remains challenging due to the unpredictability of seizure events. An accurate forecast of seizure onset helps to reduce risks in epilepsy patients. In this paper, we propose EEG-FuseFormer, a transformer-based feature fusion framework for seizure-onset prediction that combines intermediate features extracted from Convolutional Neural Networks-Long Short-Term Memory (CNN–LSTM) and ResNet-18 networks. The CNN-LSTM architecture captures both spatial and temporal features directly from the raw signal, whereas the ResNet-18 extracts features from the Short-Time Fourier Transform (STFT) representation of the EEG signals. Fusion is carried out using a transformer encoder, and the final prediction is generated using fully connected dense layers. The CHB-MIT dataset was used to validate the proposed model. The results show that the proposed model achieves a mean recall of 98.85\% and outperforms most of the state-of-the-art methods. This study evaluates the ability of the proposed feature fusion model to generalize in cross-patient testing scenarios. Fine-tuning pre-trained models on limited target patient data (target adaptation) within the cross-patient validation framework results in higher recall, precision, and F1-score metrics in comparison to the conventional cross-patient validation approach. Finally, the runtime‑based computational complexity of the model is assessed across diverse hardware platforms to highlight the performance–complexity trade‑off.

\end{abstract}


\section{Introduction} \label{sec:introduction}
Epilepsy is a brain condition that is characterized by recurring seizures, affecting nearly 50 million people around the world, making it one of the most common neurological disorders in the world \cite{worldhealthorganization_2024_epilepsy}. An automated patient-specific seizure onset prediction approach could reduce the serious consequences of seizure by detecting the onset and raising an alarm. Electroencephalograms (EEG), which measure brain electrical activity, are commonly employed to diagnose seizure onset prediction \cite{ref2024}. In epileptic patients, brain activity typically consists of four phases: pre-ictal (just before a seizure), ictal (during a seizure), postictal (immediately after a seizure), and interictal (normal activity) \cite{ref2024}. The detection of pre-ictal and interictal states is the main task in seizure onset prediction. 



EEG signals are inherently complex and exhibit irregular variations over time. Moreover, pre-ictal and interictal EEG patterns differ significantly across individuals, making seizure prediction highly challenging \cite{8788635}. Feature extraction from decomposed EEG signals across time, frequency, and time–frequency domains is a key process in many seizure onset prediction studies \cite{ref_jahan}. These extracted features are then utilized to develop automated seizure prediction models. With the rapid advancement of deep learning, Convolutional Neural Networks (CNN) \cite{truong_2018_convolutionalneuralnetworks, 8788635, 9347465}, Long Short-Term Memory (LSTM) \cite{tsiouris_2018_a, wu_2023_an}, CNN-Bidirectional-LSTM \cite{daoud_2019_efficient}, Residual network (ResNet)-LSTM \cite{lee_2024_a} architectures among others have become the most popular methods for seizure prediction in the literature. A single EEG feature cannot comprehensively represent the EEG signals, leading to insufficient accuracy and reliability in the detection of epileptic seizures \cite{zhang_2024_a}. To address this, various studies fuse multiple features to further enhance detection performance \cite{JOHN2025103253, JOHN2025109901}. Tsiouris \etal \cite{tsiouris_2018_a} fused both time and frequency domain features to select the optimal feature set for classification, to further improve classification accuracy. Zhang \etal in \cite{zhang_2024_a} proposed a Convolutional Neural Network-Gated Recurrent Unit-Attention mechanism (CNN-GRU-AM) model, where EEG signals are first decomposed using wavelet transform. Time–frequency domain and nonlinear features are then extracted, followed by further feature extraction and classification using the CNN-GRU-AM model.

However, simply combining various features can significantly increase runtime and introduce irrelevant features that may impact the final results. Hence, we propose a transformer-driven feature fusion model, EEG-FuseFormer, for seizure onset prediction. The proposed model uses deep learning networks to extract spatial and temporal dynamic features from EEG data. 
A two-channel feature fusion model based on CNN-Bi-LSTM was proposed in \cite{10156816}. In this approach, CNN was used to extract spatial features, Bi-LSTM captures temporal features, and an attention module assigns weights to the features from each module. The proposed EEG-FuseFormer aims to further enhance seizure prediction performance by incorporating a transformer encoder that can carry out self-attention to fuse the features from the ResNet-18 and 1D-CNN-LSTM architectures. 


Furthermore, most studies on seizure onset prediction in the literature have concentrated on patient-specific models aimed at predicting seizures for individual patients \cite{tsiouris_2018_a, Ricardo2022, lee_2024_a, 9347465}. Seizure symptoms vary among patients, from brief interruptions in activity to severe convulsions \cite{ref2024}. The real-life use of seizure prediction methods would require the deployment of cross-patient evaluation methods, such as leave-one-patient-out validation schemes \cite{ref2024}. The ability of the proposed EEG-FuseFormer model to generalize in cross-patient testing scenarios is evaluated. The most common approach in the literature for addressing cross-patient evaluation challenges is the application of the adaptation technique \cite{rev_S}, which allows fine-tuning of pre-trained models using a limited amount of data from target patient. This paper evaluates the performance of the proposed model with and without adaptation under the cross-patient testing framework. The results show a significant improvement in recall, precision, and F1-score metrics with target adaptation in comparison to the conventional cross-patient validation approaches. Lastly, the runtime‑based computational complexity of the proposed model is evaluated across diverse hardware platforms, including GPUs such as NVIDIA L4 and NVIDIA A100 SXM4 as well as embedded devices like the NVIDIA Jetson Nano, to illustrate the performance–complexity trade‑off.

The remainder of this paper is organized as follows: Section \ref{sec:methodology} details the proposed EEG-FuseFormer Feature Fusion model and the cross-patient testing framework. Section \ref{sec:results} outlines the experimental setup and presents the results. Finally, Section \ref{sec:conclusion} concludes the paper.

\section{Methodology}
\label{sec:methodology}
\subsection{Datasets}
The CHB-MIT Scalp EEG dataset \cite{guttag_chbmit} was used in this paper, which comprises of continuous scalp EEG recordings from 23 pediatric subjects with intractable seizures. The recordings were sampled at 256Hz and used the international 10-20 system for recordings using bipolar channels. To ensure consistency across all subjects, 18 channels were selected: FP1-F7, F7-T7, T7-P7, P7-O1, FP1-F3, F3-C3, C3-P3, P3-O1, FP2-F4, F4-C4, C4-P4, P4-O2, FP2-F8, F8-T8, T8-P8-0, P8-O2, FZ-CZ, and CZ-PZ.



Seizure activity in this work is categorized according to the proximity of the signal sample to the seizure (ictal) event. The pre-ictal period refers to the 1 hour preceding a seizure, post-ictal period refers to the 30 minutes immediately following a seizure event \cite{Ricardo2022}. The inter-ictal period refers to periods outside of the two previously defined intervals and the ictal interval. 
Due to the high irregularity of post-ictal brain activity and inter-patient variability, post-ictal segments were excluded from analysis in this work. Additionally, to ensure relevance of this study to real-world applications, cases with greater than 12 seizures were omitted from analysis since such patients would require medical intervention or continuous monitoring, making seizure prediction less valuable. Hence a total of 13 patients were selected for analysis as listed in Table \ref{tab:selected-patients}.

\begin{table}
    \centering
    \caption{Subject details from the CHB-MIT Dataset}
    \label{tab:selected-patients}
    \begin{tabular}{l l l l l p{1cm}}
    \hline
    \textbf{Case} & \textbf{Age} & \textbf{No. of} & \textbf{No. of} & \textbf{Recording Duration } \\
    & (years) & \textbf{channels} & \textbf{Seizures} & (hh:mm:ss) \\
    \hline
    chb01 &  11 & 18 & 7 & 40:33:08 \\
    chb02 &  11 & 18 & 3 & 35:15:59 \\
    chb03 &  14 & 18 & 7 & 38:00:06 \\
    chb05 &  7 & 18 & 5 & 39:00:10 \\
    chb07 &  14.5 & 18 & 3 & 67:03:08 \\
    chb09 &  10 & 18 & 4 & 67:52:18 \\
    chb13 &  3 & 18 & 12 & 33:00:00 \\
    chb14 &  9 & 18 & 8 & 26:00:00 \\
    chb18 &  18 & 18 & 6 & 35:38:05 \\
    chb19 &  19 & 18 & 3 & 29:55:46 \\
    chb20 &  6 & 18 & 8 & 27:36:06 \\
    chb21 &  13 & 18 & 4 & 32:49:49 \\
    chb23 &  6 & 18 & 7 & 26:33:00 \\
    \hline
    \end{tabular}
\end{table}

\subsection{Pre-processing}\label{subsub-sec:pre-processing}

To eliminate noise sources and to focus on frequencies relevant to epilepsy analysis, the raw EEG signals were filtered using a high-pass filter with a cutoff frequency of 1 Hz and a notch filter to remove power-line interference between 50 to 60 Hz. The data were then converted to a dataframe and interictal and preictal events are labeled according to the seizure phase. The accurate classification of pre-ictal events is considered as early detection of seizure onset in this paper. The post-ictal periods were discarded from the analysis. The class window size is 5 seconds \cite{tsiouris_2018_a}. Since there exists a class imbalance problem with fewer pre-ictal samples, the use of a sliding window approach was employed to artificially augment the number of samples by creating overlapping segments \cite{truong_2018_convolutionalneuralnetworks} during the training phase. The stride length of the sliding window selected was unique to each patient and was determined based on the label distribution. Lastly, following the aforementioned over-sampling, random under-sampling was applied to further balance cases where severe imbalance was not resolved simply using data augmentation. To ensure a leakage‑free evaluation, we adopted dataset split before windowing and augmentation. 



\begin{figure}
    \centering
    \includegraphics[scale=0.43]{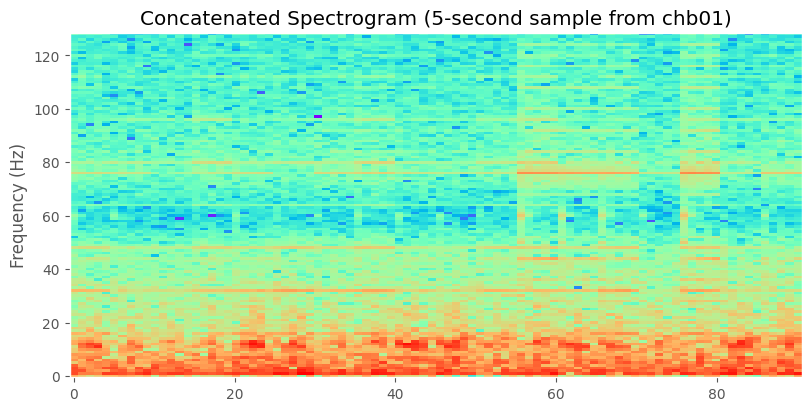}
    \caption{Sample Concatenated Spectrogram from chb01}
    \label{fig:concat-spectrogram}
\end{figure}

\subsection{Feature engineering} 
Short-Time Fourier Transform (STFT) was used to extract time-frequency features from a short-duration segment of EEG signal by computing its Fourier Transform. This results in the creation of
spectrograms that describe the power distributions of frequency over a time segment. STFT was applied to 5-second segments of EEG to extract a spectrogram representation of the sample. These were used as image inputs to Convolutional Neural Networks (CNNs) in order to extract features or classify signals. Lastly, the spectrograms from the eighteen channels were concatenated to generate the input representation for these networks. Fig. \ref{fig:concat-spectrogram} shows the concatenated spectrogram of a 5-second sample from chb01.


\subsection{EEG-FuseFormer model}
\begin{figure*}
    \centering
    \includegraphics[scale=0.5]{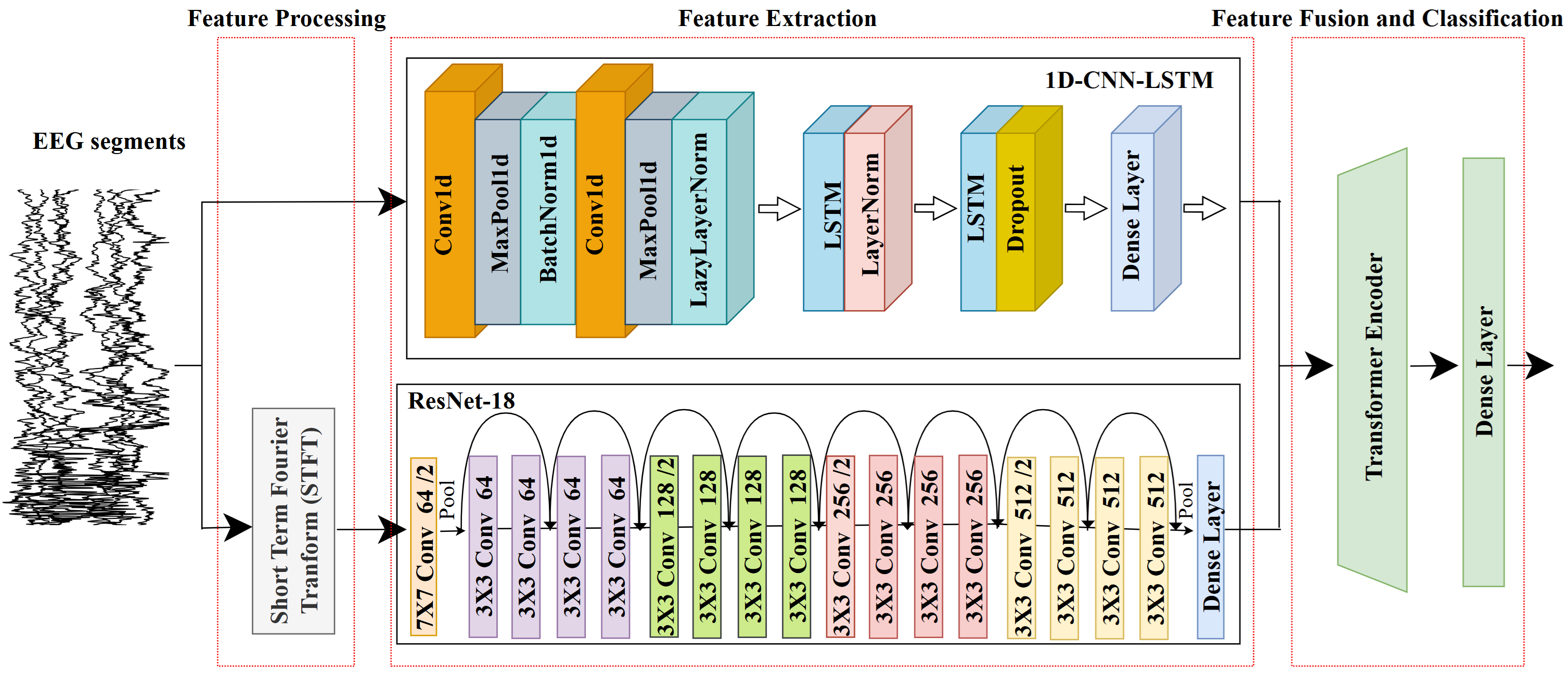}
    \caption{Proposed EEG-FuseFormer Feature-Fusion model}
    \label{fig:feature-fusion-model}
\end{figure*}

In this paper, we propose a fusion model based on ResNet-18 and 1D-CNN-LSTM for feature extraction, feature selection, and classification of epileptic EEG signals, as illustrated in Fig. \ref{fig:feature-fusion-model}. The model consists of two parallel channels that process EEG signals simultaneously. Channel 1 operates on the filtered time-domain signal and employs a 1D-CNN-LSTM architecture to capture both spatial and temporal features. Channel 2 performs frequency-domain feature engineering using an STFT block and applies ResNet-18 with ReLU activation to extract spatial features. These features are projected into a common vector dimension of size 64 using a linear layer before being connected to a multi-head transformer encoder and dense layers for final classification. The classification of interictal and preictal stages is performed using a softmax function.

\subsubsection{1D-CNN-LSTM}
The proposed model contains two convolutional blocks that act on the raw signal, each consisting of a batch normalization layer, a Rectified Linear Unit (ReLU) activation function and a max pooling layer. The first convolutional block uses a 1D convolutional layer with 18 convolutional kernels of size 7 and max-pooling with a kernel size of 5. The second block uses 128 kernels with a kernel size of 5 and a max-pooling kernel of size 7. The output of these blocks is layer-normalized and fed into a two-layer LSTM model with 128 and 64 hidden units, respectively. A ReLU activation function is used in all hidden layers. Lastly, to prevent overfitting, batch normalization was applied before convolutional layers, while layer normalization was applied before recurrent layers \cite{ba_2016_layer}. A dropout probability of 0.3 was also used along with L2-regularization with a coefficient of 0.00001 in the 1D-CNN-LSTM model. Across the models, we employed the categorical cross entropy loss function in conjunction with the Adam optimizer to reduce the need for extensive hyperparameter tuning. 


\subsubsection{ResNet-18}
ResNets improve upon the traditional CNN architecture due to the addition of residual/skip connections, which enable deeper architectures without performance degradation \cite{he_2015_deep}. ResNet-18  consists of 5 blocks as shown in Fig. \ref{fig:feature-fusion-model}. The input layer of ResNet-18 architecture is modified in this work to accept a concatenated spectrogram input of size $128 \times 90$. The extracted feature from ResNet-18 is converted into a 512-dimensional vector using a pooling operation followed by a flatten layer. 

\subsubsection{Transformer Encoder}
The transformer encoder integrates features from both ResNet-18 and 1D-CNN-LSTM architectures. The encoder used here consists of two layers, each containing two sub-layers: (1) a multi-head self-attention mechanism with 8 heads, and (2) a fully connected feed-forward network \cite{vaswani2023attentionneed, reftransformer, book1}. For the transformer encoder, all layers output dimension is 128, which is then passed through two dense layers with input sizes of 128 and 64 for the final classification.



\subsection{Cross-patient Testing Framework}
\begin{figure}
    \centering
    \includegraphics[scale=0.35]{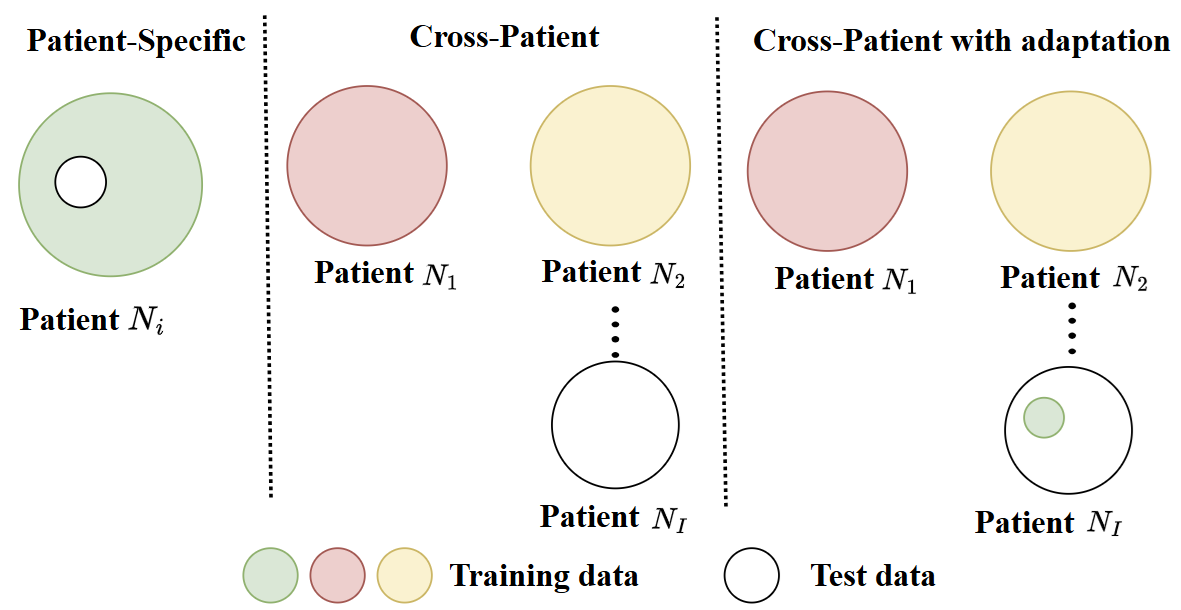}
    \caption{Visualization of the Testing Frameworks}
    \label{fig:generalized-testing-framework}
\end{figure}

The evaluation of seizure prediction models can be conducted using several testing frameworks, as shown in Fig. \ref{fig:generalized-testing-framework} \cite{ref2024}. Patient-specific modelling involves using data from a single patient to train a specific model for that patient. However, this model does not generalize to seizure onset prediction for unseen patients. To address this, we employed cross-subject modelling, where labeled data from $ \left(  N_{1}, N_{2}.. N_{I-1} \right)$ patients were used for training, and data from $ N_{I} $ patient was reserved for testing. The study investigates adaptation method to improve the performance of cross-subjects modelling approach. The target adaptation helps to finetune pre-trained models on a small sample of data from the $ N_{I} $ patient data for 3-5 epochs before testing. Target adaptation allows good classification results in cross-subjects modelling approach. These approaches aim to learn a set of relevant features from the general population that are subsequently adapted to the clinical data from the target patients. 



\section{Results and Discussion}
\label{sec:results}
The section discusses the results of proposed models when evaluated on the thirteen chosen cases in the CHB-MIT dataset. Pytorch and Torchvision were used to implement the proposed architecture. The models were trained on a 60-15-25 training-validation-test split ratio.

\begin{table*}
    \centering
    \caption{Patient-specific results for 1D-CNN-LSTM, ResNet-18 models and EEG‑FuseFormer. P, R and FAR denote Precision, Recall and False Alarm Rate respectively.}
    \begin{tabular}{l| l l l l| l l l l| l l l l}
    \toprule
    \textbf{Case} & \multicolumn{4}{c|}{\textbf{1D-CNN-LSTM}} & \multicolumn{4}{c|}{\textbf{ResNet-18}} & \multicolumn{4}{c}{\textbf{EEG‑FuseFormer}} \\
    \cline{2-13}
    & P(\%)&R(\%) &F1-Score(\%)&FAR(/h) &P(\%)&R(\%) &F1-Score(\%)&FAR(/h)& P(\%)&R(\%) &F1-Score(\%)&FAR(/h)\\
    \midrule
chb01&	91	&99	&94	&0.26	&97	&98	&97	&0.2	&97&	99	&98&	0.1 \\
chb02&	94&	98	&95	&0.25	&99	&99	&98&	0.19&	99&	98&	99 &	0.13 \\
chb03&	95&	100	&97	&0.14	&96	&99	&99	&0.14&	98	&99	&98	&0.06 \\
chb05&	81&	90	&85	&0.24	&93	&95	&94&0.2	&94	&98	&96	&0.1 \\
chb07&	94&	99	&96	&0.23	&98	&100	&98	&0.18	&99	&100	&100	&0.089 \\
chb09&	87&	96	&91	&0.22	&96	&97	&96	&0.17	&96	&99	&98	&0.109 \\
chb13&	88&	96	&91	&0.24	&97	&99	&98	&0.1	&97	&100	&100	&0.06 \\
chb14&	72&	80	&75	&0.27	&91	&94	&92	&0.18	&95	&97	&94	&0.14 \\
chb18&	98&	100	&99	&0.13	&98	&99	&99	&0.09	&99	&100	&99	&0.03 \\
chb19&	100	&100	&99	&0.14	&98	&99	&99	&0.06	&99	&100	&100	&0.06 \\
chb20&	95&	99	&97	&0.16	&98	&98	&98	&0.09	&98	&100	&100	&0.04 \\
chb21&	84&	97	&89	&0.24	&95	&93	&94	&0.20	&95	&96	&95	&0.13 \\
chb23&	75&	61	&67	&0.50	&96	&98	&97	&0.32	&97	&99	&96	&0.15 \\
    \bottomrule
    \end{tabular}
    \label{tab:1d-cnn-results}
\end{table*}

Most of the works reported in the literature report patient-specific results, due to the individualized nature of seizures (i.e., models were trained and tested on the same patient's data). In order to ensure a consistent comparison, the models were first tested using the same patient-specific setup. Table \ref{tab:1d-cnn-results} shows the results on individual cases with 1D-CNN-LSTM, ResNet-18 and EEG‑FuseFormer models. Precision, Recall, F1-Score and False Alarm Rate (FAR) are the performance metrics employed for the evaluation. Precision quantifies the proportion of predicted positive events that correspond to true positive outcomes. Recall denotes the percentage of actual seizure events that are correctly identified by the predictive model \cite{8788635}. FAR quantifies how many inter‑ictal samples are misclassified as pre‑ictal per hour of inter‑ictal activity \cite{8788635}. The results in Table \ref{tab:1d-cnn-results} show that the 1D-CNN-LSTM model achieves a mean recall of 93.4\% and a mean F1-score of 90.38\%.In comparison, ResNet-18 demonstartes a superior performance achieving a mean recall of 97.53\% and a mean F1-Score of 96.8\%. The EEG-FuseFormer has the highest predictive performance, with a mean recall of 98.85\% and a mean F1-Score of 97.92\%. Additionally, EEG-FuseFormer shows a low FAR(/h) of 0.12, whereas the mean FAR(/h) is at 0.27 for 1D-CNN-LSTM.

\begin{table*}
    \centering
    \caption{Comparison of the proposed EEG-FuseFormer model with the state-of-the-arts. NA: Not Available}
    \begin{tabular}{l| p{6cm}|  p{4cm}| p{1.5cm} |p{1cm}| p{1cm} }
    \toprule
    \textbf{Ref} & \textbf{Features} & \textbf{Classifier} & \textbf{Preictal duration (min)}&\textbf{Recall (\%)} & \textbf{FAR(/h)} \\
    \midrule
    Truong \etal \cite{truong_2018_convolutionalneuralnetworks} & STFT &CNN & 5&81.20 & 0.16\\
    Tsiouris \etal \cite{tsiouris_2018_a} & Time-domain: Statistical moments, Standard Deviation, Peak-to-peak voltage Frequency-domain: Fast Fourier Transform, Power Spectral Density, Discrete Wavelet Transform  &LSTM & 60&99 & 0.06\\
    Zhang \etal \cite{8788635} &Spatial pattern statistics & CNN& 30&92.2 & 0.12\\
     Jemal \etal \cite{9777979} & Raw Data & CNN inspired by Filter Bank
Common Spatial Pattern (FBCSP)& 30&96.1  & 0.04\\
    Wu \etal \cite{wu_2023_an} & Raw Data &LSTM & 60&92.17 & 0.29\\
     Ma \etal \cite{10156816} & Raw Data &CNN-Bi-LSTM & NA &94.84 & NA\\
    Lee \etal \cite{lee_2024_a} & STFT &ResNet-LSTM & 30& 92.71 & 0.05\\
    Zhang \etal \cite{zhang_2024_a} & Time-frequency domain with non-linear fusion &CNN-GRU-AM & NA &95.47 & NA\\
    \textbf{This Work} & Raw Data&1D-CNN-LSTM & 60&93.46 & 0.23\\
    \textbf{This Work} & STFT&ResNet-18 & 60&97.53 & 0.15\\
    \textbf{This Work} & Raw Data and STFT fusion &EEG-FuseFormer & 60&98.85 & 0.09\\
    \bottomrule
    \end{tabular}
    \label{tab:results-summary}
\end{table*}


Table \ref{tab:results-summary} shows the results of the proposed feature fusion model and its comparison with the state-of-the-art models. The results indicate that feature fusion models by integrating multiple features outperform standalone models like 1D-CNN-LSTM and ResNet-18. The proposed fusion model outperforms the state‑of‑the‑art methods of Truong \etal \cite{truong_2018_convolutionalneuralnetworks}, Zhang \etal \cite{8788635}, Jemal \etal \cite{9777979}, Wu \etal \cite{wu_2023_an}, Ma \etal \cite{10156816}, Lee \etal \cite{lee_2024_a} and Zhang \etal \cite{zhang_2024_a}. Additionally, the models in \cite{tsiouris_2018_a} require expert knowledge on the appropriate use of features due to the elaborate extraction of features such as entropy characteristics, peak-to-peak voltage, etc. The proposed model achieves comparable performance with Tsiouris \etal \cite{tsiouris_2018_a} but with significantly lower feature engineering efforts. 


\subsection{Cross-patient Testing Framework}\label{sub-sec:cp-framework}
Table \ref{tab:cp-testing-no-finetuning} shows the results of the proposed models for the cross-patient testing framework with and without adaptation. In the cross-patient testing, data from 12 patients were used for model training, while the remaining patient data was used for target adaptation and testing. Here, chb023 was the target patient for adaptation and testing. Observing the results in Table \ref{tab:cp-testing-no-finetuning}, it is apparent that a major limitation of these models is their patient-specificity. However, after finetuning the training model with the target adaptation technique, we can improve the model performance. This is evident from the significant improvement in the recall values, which increased from $40-60$\% across different models. Furthermore, Table \ref{tab:cp-testing-no-finetuning} clearly demonstrates the superior performance of the EEG-FuseFormer model compared to CNN-LSTM and ResNet-18. The proposed model achieves a 22\% improvement over 1D-CNN-LSTM and a 4\% improvement over ResNet-18 within the cross-patient testing framework with target adaptation.

\subsection{Computational Complexity Analysis}
Table \ref{runtime} shows the computational complexity analysis of the three models CNN‑LSTM, ResNet‑18, and EEG‑FuseFormer based on runtime performance. Our evaluation utilizes multiple GPU platforms, from NVIDIA L4 and NVIDIA A100 SXM4 to embedded devices like the Jetson Nano. To assess computational complexity, we compared the CNN‑LSTM, ResNet‑18, and EEG‑FuseFormer architectures in terms of both runtime and the number of trainable parameters. Table \ref{runtime} shows the mean runtime for patient‑specific testing of the subjects listed in Table \ref{tab:selected-patients}. To evaluate runtime on a memory-constrained device such as the Jetson Nano, the data were divided into 10-hour segments and processed sequentially to estimate computational complexity. In real-time operation, this segmentation will not be necessary, as seizure-detection data will be streamed continuously, processed on the fly, and discarded after prediction. 

Based on the results in Table \ref{tab:1d-cnn-results} and Table \ref{runtime}, EEG‑FuseFormer delivers the highest predictive performance by leveraging multi‑scale feature fusion and attention mechanisms. However, this gain comes with increased architectural complexity and runtime overhead. In contrast, CNN‑LSTM model achieves competitive performance with relatively low complexity, making it suitable for resource-constrained deployment. Thus, selecting an appropriate model depends on whether the target application is cloud-based or edge deployment. Overall, the results highlight a clear performance and complexity trade‑off across CNN‑LSTM, ResNet‑18, and EEG‑FuseFormer.



\begin{table}
    \centering
    \caption{Cross-patient testing with and without adaptation technique.}
    \begin{tabular}{p{1.8cm}| p{1.6cm} |p{1cm} |p{1.2cm} |p{1.1cm}}
    \toprule
    \textbf{Ref} & \textbf{Model} & \textbf{Recall (\%)} & \textbf{Precision (\%)} & \textbf{F1-score(\%)} \\
    \hline
    \multicolumn{5}{l}{{Cross-patient testing without adaptation}}\\
    \hline
    Jemal \etal \cite{ref2024} & CNN & - & - & 55.34\\
    \textbf{This work} & {1D-CNN-LSTM} & 54 & 40  & 46  \\
    \textbf{This work} & {ResNet-18} & 59  & 56 & 57  \\
    \textbf{This work} & {EEG-FuseFormer} & 59  & 59 & 58 \\
    \hline
    \multicolumn{5}{l}{{Cross-patient testing with adaptation}}\\
    \hline
    Jemal \etal \cite{ref2024} & CNN & - & - & 66.45\\
    Deng \etal \cite{10156816} & CNN & - & - & 75 \\
    \textbf{This work} & {1D-CNN-LSTM} & 76 & 67  & 66  \\
    \textbf{This work} & {ResNet-18} & 89  & 84  & 86  \\
    \textbf{This work} &  {EEG-FuseFormer} & 93  & 82  & 88  \\
    \bottomrule
    \end{tabular}
    \label{tab:cp-testing-no-finetuning}
\end{table}

\begin{table}
    \centering
    \caption{Computational Complexity in terms of runtime and the number of trainable parameters}
\begin{tabular}{p{1.5cm}|p{1cm} |p{1.5cm} |p{0.8cm}|l}
\toprule
     \multirow{2}{*}{Model}     & \multicolumn{3}{c|}{Runtime (seconds)}                                                      &Trainable \\ \cline{2-4}
          & NVIDIA L4 & NVIDIA A100 SXM4 & Jetson Nano & Parameters                                      \\ \midrule
CNN-LSTM  & \multicolumn{1}{l|}{0.50K}    & \multicolumn{1}{l|}{0.20K}           &      1.58K              &   0.12M                                                         \\ 
ResNet-18 & \multicolumn{1}{l|}{3.28K}   & \multicolumn{1}{l|}{1.23K}          &       8.18K             &    11.17M                                                        \\ 
EEG‑ FuseFormer    & \multicolumn{1}{l|}{22.47K}  & \multicolumn{1}{l|}{11.29K}         &  56.17K                  &   12.40M                                                         \\ \bottomrule
\end{tabular}
\label{runtime}
\end{table}



\section{Conclusion}
\label{sec:conclusion}
The paper investigates the use of feature fusion deep learning models for seizure onset prediction. 1D-CNN-LSTM and ResNet-18 architectures were explored that leverage both time- and frequency-domain features from EEG signals. ResNet-18 appeared to be the best performing model with a mean sensitivity of 97.9\%. The paper proposed EEG-FuseFormer, a novel feature-fusion framework combining the outputs of both the ResNet-18 model and the 1D-CNN-LSTM model, exploiting the self-attention framework of transformer encoders. The proposed fusion model was compared to existing state-of-the-art models and demonstrated improved performance. To better assess the ability of the proposed model to generalize to new patients, a cross-patient modelling approach was employed. The adaptation techniques enhanced the performance of the cross-patient testing framework by over 40\%. The predictive performance and complexity analysis shows that
there should be a trade-off across CNN-LSTM, ResNet-18, and EEG-FuseFormer in selection of different seizure prediction target applications. The results of the complexity analysis show that selecting CNN‑LSTM, ResNet‑18, or EEG‑FuseFormer models involves navigating a trade‑off between performance and complexity based on the target seizure prediction application.


\vspace{10pt}
\section{Acknowledgements}
This project received funding from the European Union’s Horizon Europe Excellent Science programme under the Marie Skłodowska-Curie Actions Grant Agreement [Grant Agreement No 101081457] and in part from CÚRAM, Science Foundation Ireland Research Centre under grant number 13/RC/2073\_P2, as well as from Research Ireland-WIN Research Alliance Award. We also acknowledge the support and computational facilities provided by the Irish Centre for High-End Computing (ICHEC).

\vspace{10pt}
\balance
\bibliographystyle{IEEEtran}
\bibliography{ref.bib}

\end{document}